\documentclass[10pt,twocolumn,letterpaper]{article}

\usepackage[pagenumbers]{cvpr} 

\usepackage[accsupp]{axessibility}

\definecolor{cvprblue}{rgb}{0.21,0.49,0.74}
\usepackage[pagebackref,breaklinks,colorlinks,allcolors=cvprblue]{hyperref}

\def\paperID{46334} 
\def\confName{CVPR}
\def\confYear{2026}

\title{OVOD-Agent: A Markov-Bandit Framework for Proactive Visual Reasoning and Self-Evolving Detection}

\author{
Chujie Wang\textsuperscript{*} \quad
Jianyu Lu\textsuperscript{*} \quad
Zhiyuan Luo \quad
Xi Chen\textsuperscript{$\dagger$} \quad
Chu He\textsuperscript{$\dagger$} \\
Wuhan University \\
{\tt\small chujie@whu.edu.cn}
}

\usepackage[ruled,vlined,linesnumbered]{algorithm2e}
\DontPrintSemicolon
\SetKwInOut{Input}{Input}
\SetKwInOut{Output}{Output}
\SetKwInOut{Params}{Params}

\makeatletter
\renewcommand\paragraph{\@startsection{paragraph}{4}{\z@}%
  {1.5ex \@plus 0.2ex \@minus 0.1ex}%
  {-0.6em}%
  {\normalfont\normalsize\bfseries}}
\makeatother

\usepackage{tabularx}
\usepackage{booktabs}
\usepackage[table]{xcolor}
\usepackage{multirow}

\usepackage{multicol}
\usepackage{stfloats}

\begin{document}
\maketitle
\begin{abstract}
Open-Vocabulary Object Detection (OVOD) aims to enable detectors to generalize across categories by leveraging semantic information. Although existing methods are pretrained on large vision-language datasets, their inference is still limited to fixed category names, creating a gap between multimodal training and unimodal inference. Previous work has shown that improving textual representation can significantly enhance OVOD performance, indicating that the textual space is still underexplored. To this end, we propose OVOD-Agent, which transforms passive category matching into proactive visual reasoning and self-evolving detection. Inspired by the Chain-of-Thought (CoT) paradigm, OVOD-Agent extends the textual optimization process into an interpretable Visual-CoT with explicit actions. OVOD's lightweight nature makes LLM-based management unsuitable; instead, we model visual context transitions as a Weakly Markovian Decision Process (w-MDP) over eight state spaces, which naturally represents the agent's state, memory, and interaction dynamics. A Bandit module generates exploration signals under limited supervision, helping the agent focus on uncertain regions and adapt its detection policy. We further integrate Markov transition matrices with Bandit trajectories for weakly-supervised Reward Model (RM) optimization, forming a closed loop from Bandit exploration to RM learning. Experiments on COCO and LVIS show that OVOD-Agent provides consistent improvements across OVOD backbones, particularly on rare categories, confirming the effectiveness of the proposed framework.
\end{abstract}    
\section{Introduction}
\label{sec:intro}

\begin{figure}[t]
\centering
\includegraphics[width=1.0\columnwidth]{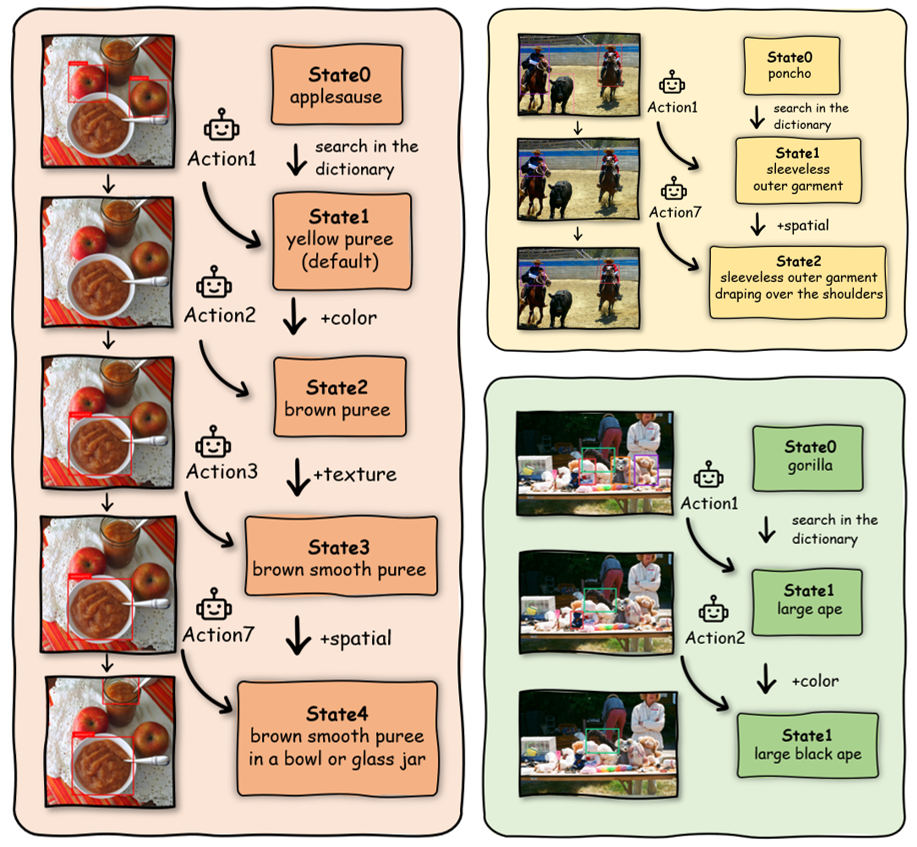} 
\caption{We illustrate the state-transition behavior of OVOD-Agent as it iteratively updates its category hypothesis. Starting from an initial dictionary lookup, the agent applies attribute-aware actions that adjust color, texture, and spatial cues to produce a more accurate and grounded state description. The number of required actions varies across images, from single-step updates to multi-step reasoning.}
\label{fig1}
\vspace{-3mm}
\end{figure}

Open-Vocabulary Object Detection (OVOD) aims to extend object detectors to arbitrary concepts by exploiting the semantic priors learned from large-scale vision–language pretraining~\cite{Detic,glip,clip,ovd}. With improved region–text alignment and large-vocabulary modeling~\cite{gdino,yoloworld,detclip,detclipv2,detclipv3,dinox}, recent methods have substantially enhanced open-set recognition. However, although these models are trained with multimodal supervision, their inference still depends on a fixed set of category names. This turns detection into a simple matching routine and creates a clear mismatch between multimodal training and unimodal inference. Consequently, existing methods struggle to reason under visual ambiguity, adapt to unfamiliar contexts, and detect rare or fine-grained categories.

A growing body of work has shown that the textual space has a much greater impact on OVOD performance than previously assumed. Techniques such as prompt learning~\cite{promptovod}, prompt diversification~\cite{zsmm}, attribute-based descriptions~\cite{classification}, and automatic class-name optimization~\cite{huawei} consistently reshape the textual embedding space and lead to marked improvements across OVOD benchmarks, suggesting that this space remains far from saturated~\cite{fg}. To enrich fine-grained semantics, recent studies introduce LLM-generated priors to expand the textual domain~\cite{llmguided1,llmguided2}. Yet these approaches remain essentially static: a single adjustment to the descriptor cannot capture the evolving relationships among regions, contexts, and attributes during detection. What is still lacking is a context-dependent, iterative mechanism that can continuously refine textual representations throughout inference.

The step-by-step reasoning paradigm of CoT~\cite{cot}, together with advances in interactive vision–language agents~\cite{agentsurvey,react,toolformer}, has enabled multi-step visual operations in multimodal systems. Frameworks such as Visual-CoT~\cite{vcot,vocot} suggest that visual cues can be aligned with textual semantics through progressive reasoning. In OVOD, CoT-PL~\cite{cotpl} further shows that incorporating such reasoning during training can improve the quality of pseudo labels. However, these approaches inherit the agent-style design that places an LLM at the center of the decision process~\cite{cotpl,llmguided2}, which imposes substantial computational and memory overhead during inference. Some even rely on multiple rounds of human feedback to guide the reasoning procedure~\cite{humanfeedback}. Such designs run counter to the core strengths of object detection—speed, scalability, and ease of deployment—and ultimately make the pipeline unnecessarily heavy for the OVOD setting.

To address these limitations, we introduce OVOD-Agent, a lightweight and LLM-free framework that transforms OVOD from passive category matching into proactive visual reasoning and self-evolving detection. We model the evolution of visual and semantic cues as a Weakly Markovian Decision Process (w-MDP) defined over eight compact visual states, which provides a structured view of state transitions and memory and serves as the foundation for our agent formulation. A Bandit-based exploration module identifies uncertain or semantically ambiguous regions, generating informative trajectories for adaptive refinement. We further couple Markov transition statistics with these trajectories to train a weakly-supervised Reward Model (RM), forming a closed loop that enables continuous policy improvement under weak supervision. Experiments on COCO and LVIS show that OVOD-Agent consistently enhances existing OVOD backbones while adding only limited deployment overhead (\textless 100 MB disk, \textless 20 MB memory) and \textless 100 ms latency cost, demonstrating its practicality for real-world deployment. In summary, the main contributions are as follows:
\begin{itemize}
    \item We introduce OVOD-Agent, which models visual context transitions using an eight-state weak MDP and converts static prompt matching into an interpretable multi-step Visual-CoT reasoning process with explicit actions, interaction, and memory.
    \item We propose a Bandit-based exploration strategy for collecting trajectories under uncertain visual states and combine global Markov transition statistics with offline trajectories to train a weakly-supervised Reward Model, forming a closed-loop mechanism for agent self-evolution.
    \item Our framework introduces no large-model dependencies, incurs minimal inference overhead, preserves the deployment efficiency of existing detectors, and remains compatible with a wide range of OVOD backbones.
\end{itemize}
\section{Related Work}
\label{sec:related}

\paragraph{LLM-Based Textual Optimization}
Large language models (LLMs) have recently been adopted to enhance textual representations in vision--language tasks~\cite{minigpt4,kosmos2,llava,blip2}. In OVOD, LLMs are increasingly used to enrich category descriptors, generate fine-grained attributes, or provide contextual priors~\cite{llmguided1,llmguided2}. Such approaches substantially expand the semantic coverage of textual embeddings and often improve zero-shot recognition performance. However, LLM-centric frameworks typically introduce heavy computational and storage overhead due to their large parameter scales, and many require multi-round human feedback or instruction tuning~\cite{humanfeedback}. These characteristics contradict the long-standing strengths of object detectors---efficiency, scalability, and deployability. Our work differs by avoiding LLM dependence entirely and focusing on lightweight, inference-time refinement.

\paragraph{Lightweight Textual Enhancement and Discrete Alignment}
Retrieval-augmented generation (RAG)~\cite{rag} offers a simple and effective way to enrich textual descriptions by retrieving semantically relevant information from large corpora. In OVOD, however, generating high-quality retrieval vectors from limited visual evidence remains a significant challenge. RALF~\cite{ralf} addresses this issue by training an OVOD-specific retrieval module and further augments textual prompts with generative completion~\cite{diffusion}, effectively mitigating the lack of textual diversity. Yet these improvements remain fundamentally single-step and cannot support iterative, state-dependent reasoning, making it difficult to emulate the multi-step reasoning patterns observed in LLM-style Chain-of-Thought~\cite{react,toolformer,agentsurvey}. Several alignment-based approaches~\cite{regionclip,denseclip} demonstrate that region--text matching exhibits strong discreteness, where small perturbations in the textual space can lead to substantial shifts in detection behavior. This observation provides an important insight: coarse-to-fine, discrete transitions in the semantic space can act as a structured alternative to the continuous reasoning typically associated with LLM-based CoT. Markov-based formulations~\cite{markovcot,markov} further support this viewpoint by showing that discrete semantic transitions can approximate complex reasoning procedures typically attributed to LLMs. Our work builds on this line of thought by treating OVOD inference as a structured decision process. Instead of continuous textual manipulation or heuristic prompt engineering, we adopt a Weakly Markovian Decision Process (w-MDP) to model discrete visual--semantic transitions within a compact state space.

\paragraph{Sequential Decision Making and Bandit Exploration}
Reinforcement learning (RL) has been widely explored for sequential decision-making in vision systems~\cite{sequencerl,dreamerv2,rlhf,pairwise}. However, full RL pipelines are impractical in OVOD due to limited supervision and the high cost of rollout-based training. Lightweight exploration methods such as multi-armed and contextual Bandits~\cite{ubc,conbandit} offer a more suitable alternative, providing uncertainty-driven sampling without the overhead of policy learning. In our approach, Bandit exploration identifies uncertain or semantically ambiguous states and supplies informative transitions for downstream reward estimation. Combined with Markov transition statistics, this yields a compact and efficient Markov–Bandit reinforcement mechanism tailored to OVOD.
\section{Method}
\label{sec:method}

\begin{figure*}[ht]
\centering
\includegraphics[width=1.0\textwidth]{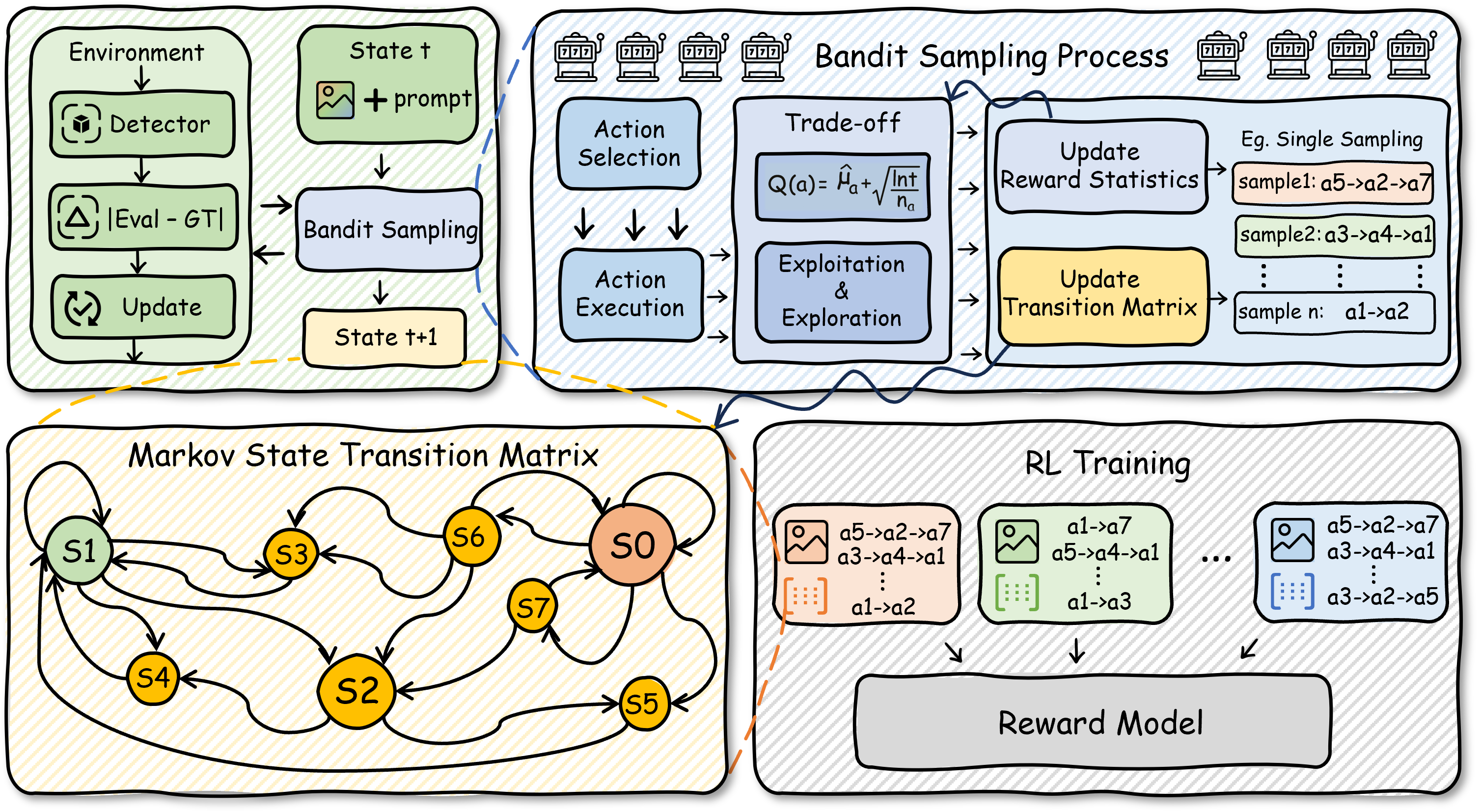} 
\caption{OVOD-Agent operates through a self-evolving visual reasoning pipeline. \textbf{(a)} The environment updates the visual state by applying detector feedback and the current prompt-conditioned context. \textbf{(b)} A UCB-based Bandit module selects and executes visual actions, collecting rewards and empirical transitions from sampled trajectories. \textbf{(c)} The collected trajectories are joined into an image-specific Markov transition matrix that models how weak Markov units evolve under different visual operations and serves as a structured prior for learning. \textbf{(d)} A lightweight Reward--Policy Model (RM) is trained on these trajectories and transition priors, distilling both transition behavior and weak reward signals. During inference, the RM replaces Bandit exploration and directly guides the agent's step-by-step refinement process without relying on LLMs.}
\label{fig2}
\end{figure*}

In this section, we introduce OVOD-Agent, a lightweight self-evolving visual agent for open-vocabulary object detection, as illustrated in Fig.~\ref{fig2}. We first present the problem formulation and the agent foundation in Sec.~\ref{sec:3.1}-\ref{sec:3.2}, followed by the data sampling procedure in Sec.~\ref{sec:3.3}, and finally the training strategy and full evolution loop in Sec.~\ref{sec:3.4}-\ref{sec:3.5}.

\subsection{Problem Setup and Notation}
\label{sec:3.1}

\paragraph{Core Idea.}
Traditional OVOD systems rely on static one-shot matching between text and visual regions, lacking the ability to jointly reason about and adjust the matching space. We aim to establish a proactive visual reasoning paradigm, where the detector actively executes a sequence of explicit visual operations under the current visual state to iteratively refine its textual representation, forming an interpretable \emph{Visual Chain-of-Thought} (Visual-CoT).

\paragraph{Formulation.}
Given an image $x$ and an open-vocabulary text prompt $T$, a detector $D$ outputs region proposals and category scores:
\begin{equation}
p = D(x, T).
\end{equation}
To introduce reasoning ability, we define a context $c_t = (x_t, T_t)$ and execute an explicit visual action $a_t \in \mathcal{A}$ at each step:
\begin{equation}
c_{t+1} = f(c_t, a_t), \quad t = 0, 1, \dots
\end{equation}
where $f(\cdot)$ denotes a deterministic or stochastic transformation over the visual context, such as color cue extraction, ROI-based context adjustment, or texture analysis. This context-driven interaction process defines our agent prototype, \textbf{OVOD-Agent}.

\paragraph{Action Space.}
We define seven interpretable primitive visual operations that constitute the agent's visual reasoning language:

\begin{table}[ht]
  \caption{Seven interpretable visual operators forming the agent's reasoning language.}
  \label{tab1}
  \centering
  \small
  \setlength{\tabcolsep}{6pt}
  \renewcommand{\arraystretch}{1.15}
  \begin{tabular}{@{}c l p{5.2cm}@{}}
    \toprule
    \textbf{ID} & \textbf{Action} & \textbf{Explanation} \\
    \midrule
    $a_1$ & Dictionary  & Alias/backoff; synonyms and hypernyms \\
    $a_2$ & Color       & HSV / cluster-based visual color cues \\
    $a_3$ & Texture     & Texture analysis (LBP/GLCM) \\
    $a_4$ & Background  & FG/BG context analysis; ROI adjustment \\
    $a_5$ & Geometry    & Geometric properties (scale, aspect ratio) \\
    $a_6$ & Lighting    & Illumination/shadow analysis (HSV-V) \\
    $a_7$ & Spatial     & Spatial relation analysis (position, IoU) \\
    \bottomrule
  \end{tabular}
\end{table}

Each operator represents one step of visual reasoning, providing a lightweight and interpretable cue within the agent's visual context.

\subsection{Weak Markovian Modeling}
\label{sec:3.2}

\paragraph{Weak Markov Decision Process (w-MDP).}
A Markov Decision Process (MDP) models decision-making based solely on the current state, naturally fitting agent design by describing the evolution of state, action, and memory. In conventional Markov decision settings, the state $s_t$ and action $a_t$ are strictly separated, with the transition dynamics defined as:
\begin{equation}
P(s_{t+1} \mid s_t, a_t).
\end{equation}
However, in proactive visual reasoning, this separation becomes unnecessary. Visual actions—based on color, texture, geometric, lighting, or spatial cues—directly update the image-text context and inherently determine the next state. Moreover, the contextual evolution of these weak units is further recorded during Bandit-driven sampling in Sec.~\ref{sec:3.3}. We therefore propose a \textbf{Weakly Markovian Decision Process (w-MDP)}, where state and action are unified into a single weak Markov unit:
\begin{equation}
z_t = g(c_t, a_t) \in \mathcal{Z},
\end{equation}
with $c_t=(x_t, T_t)$ denoting the current context and $a_t$ the visual operation.  
This unified representation captures both the contextual semantics and the applied operation, reducing the agent's context burden and forming a compact weak Markov unit on which the transition is defined.

The transition is defined as:
\begin{equation}
P(z_{t+1} \mid z_t),
\end{equation}
under a short-term memory assumption:
\begin{equation}
P(z_{t+1} \mid z_t, z_{t-1}, \dots) \approx P(z_{t+1} \mid z_t).
\end{equation}
This unified formulation preserves interpretability and avoids enumerating explicit state-action pairs, enabling lightweight transition updates. Thus, w-MDP provides a coherent framework for subsequent Bandit exploration and reinforcement distillation—each weak unit is both the result of reasoning and the starting point for the next step.

\paragraph{Base Markov Field Initialization.}
Under limited supervision, we initialize a base weak-Markov structure on the weak unit $z_0$, ensuring that both rewards and transitions are properly regularized to avoid early-stage random exploration.

\textbf{Reward Baseline (GT-seeded).}
A weak reward is defined from the mismatch between predicted and ground-truth bounding boxes:
\begin{equation}
r_t^{GT} = 1 - \mathrm{IoU}(b_t^{pred}, b_t^{GT}),
\end{equation}
which serves as a quality baseline for each $z_t$. A higher $r_t^{GT}$ indicates greater uncertainty, implying that the current state requires further refinement.

\textbf{Transition Prior (Dirichlet).}
When data are scarce, each weak unit is assigned an outgoing prior distribution:
\begin{equation}
\hat P(\cdot \mid z_t) \leftarrow \mathrm{Dirichlet}(\mathbf{n}_{z_t}),
\end{equation}
where $\mathbf{n}_{z_t}$ denotes pseudo-counts of candidate actions, typically initialized as a uniform vector. This prior guarantees probability normalization and structural feasibility for subsequent updates.

These two components jointly form the \textbf{Base Markov Field}: the GT-based reward provides weak supervision at the reward level, while the Dirichlet prior imposes structural regularization at the transition level. Together, they stabilize Bandit-driven exploration and guide early reasoning trajectories toward meaningful regions of the state space.

\subsection{Bandit-Based Exploration Strategy}
\label{sec:3.3}

\paragraph{Exploration Motivation.}
Instead of pursuing a deterministic optimal solution per image, OVOD-Agent aims to sample diverse and high-quality reasoning trajectories for training. Random exploration is inefficient, while greedy selection easily converges to local optima. We adopt a \textbf{UCB-based contextual Bandit} strategy to balance exploration and exploitation.

\paragraph{Action Selection.}
For each context $c_t$, the local mean reward and visit count are denoted as $\hat{\mu}_t(a\mid c_t)$ and $n_t(a\mid c_t)$.  
The decision rule is:
\begin{equation}
Q_t(a) = \hat{\mu}_t(a\mid c_t) + \lambda \sqrt{\frac{\ln t}{1 + n_t(a\mid c_t)}},
\end{equation}
\begin{equation}
a_{t+1} = \arg\max_a Q_t(a),
\end{equation}
where $\lambda$ controls exploration strength.  
After executing $a_{t+1}$, the transition distribution is updated via a Dirichlet prior:
\begin{equation}
\hat{P}(\cdot\mid z_t) \leftarrow \mathrm{Dirichlet}(\mathbf{n}_{z_t} + \mathbf{e}_{z_{t+1}}).
\end{equation}
This update maintains Markov consistency and stabilizes exploration.

\paragraph{Stopping Criteria.}
Each trajectory stops when one of the following holds:
\begin{itemize}
\item State stabilization: $\|c_{t+1}-c_t\| < \delta_s$;
\item Reward convergence: $|r_{t+1}-r_t| < \delta_r$;
\item Step limit: $t \ge H_{\max} = 7$.
\end{itemize}
At the image level, sampling terminates when:
\begin{itemize}
\item Mean reward increment $\Delta\bar{r} < \epsilon_r$;
\item Transition matrix convergence $\|\hat{P}_{k+1}-\hat{P}_k\|_F < \epsilon_P$;
\item Maximum episode limit $E_{\max}=50$.
\end{itemize}
Default thresholds are set to: $\delta_s=0.02$, $\delta_r=10^{-3}$, $\epsilon_r=\epsilon_P=10^{-3}$.

\subsection{Reinforcement via Markov-Bandit Feedback}
\label{sec:3.4}

\paragraph{Trajectory Dataset.}
For each image $x_i$, the Bandit procedure in Sec.~\ref{sec:3.3} generates multiple weak Markov trajectories,
\begin{equation}
\mathcal{T}_i = \{(z_t^{(m)}, z_{t+1}^{(m)}, r_t^{(m)})\}_{t,m},
\end{equation}
together with an empirical transition prior $\hat{P}_i(\cdot \mid z)$ estimated from Dirichlet updates.
The offline trajectory dataset is then
\begin{equation}
\mathcal{D} = \{(\mathcal{T}_i, \hat{P}_i)\}_{i=1}^N .
\end{equation}

\paragraph{Reward–Policy Model (RM).}
Given the offline dataset $\mathcal{D}$, the RM is designed as a lightweight dual-head network comprising:
\begin{itemize}
    \item \textbf{Policy head} $\pi_\theta(\cdot \mid z_t)$, modeling local transition continuity;
    \item \textbf{Reward head} $\hat{r}_\theta(z_t)$, predicting the expected weak reward.
\end{itemize}

\paragraph{Joint Objective.}
The RM is trained to recover the transition behavior and reward patterns encoded in the sampled trajectories $\mathcal{T}_i$ and their corresponding transition priors $\hat{P}_i$. Its learning objective integrates three components:
\begin{equation}
\begin{aligned}
\mathcal{L}_{\mathrm{RM}}
=&~ \underbrace{
\mathbb{E}_{(z_t,z_{t+1}) \sim \mathcal{D}}
\!\left[-w_t \log \pi_\theta(z_{t+1}\mid z_t)\right]}_{\text{trajectory distillation}}  \\
&+ \beta~
\underbrace{
\mathbb{E}_{(z_t,r_t) \sim \mathcal{D}}
\!\left[(\hat{r}_\theta(z_t) - r_t)^2\right]}_{\text{reward reconstruction}} \\
&+ \gamma~
\underbrace{
\mathbb{E}_{z_t \sim \mathcal{D}}
\!\left[\mathrm{D_{KL}}\!\left(\pi_\theta(\cdot\mid z_t)\Vert \hat{P}_i(\cdot\mid z_t)\right)\right]}_{\text{Markov regularization}} .
\end{aligned}
\end{equation}
The three terms encourage the RM to (i) imitate observed transition patterns,  
(ii) reconstruct the weak reward signal, and (iii) remain aligned with the empirical Markov structure of each image.

\begin{algorithm}[t]
\caption{OVOD-Agent: Markovian Evolution}
\label{alg:ovod}
\Input{Image $x$, Prompt $T$, Detector $D$}
\Params{Actions $\mathcal{A}=\{a_1,\dots,a_7\}$; RM $(\pi_\theta,\hat r_\theta)$; 
UCB coefficient $\lambda$; thresholds $\delta_s,\delta_r,\epsilon_r,\epsilon_P$; 
limits $H_{\max},E_{\max}$}
\Output{Reasoning trajectories and final detections}

\textbf{Initialize:} $c_0 \gets (x,T)$, $a_0 \gets$ \texttt{Dictionary}; 
buffer $\mathcal{D} \gets \emptyset$\;

\For{$\text{episode}=1$ \KwTo $E_{\max}$}{
  $t \gets 0$;\quad $\mathcal{T} \gets \emptyset$;\quad initialize Dirichlet counts $\{\mathbf{n}_{z}\}$\;
  \While{not \textbf{TrajStop}$(c_t,r_{t-1},t)$}{
    $z_t \gets g(c_t,a_t)$ \tcp*{weak Markov unit}
    $a_{t+1} \gets \arg\max_{a\in\mathcal{A}}
      \Big[\hat\mu(a\mid c_t) + \lambda\sqrt{\frac{\ln\max(2,t)}{1+n(a\mid c_t)}}\Big]$\;
    $c_{t+1} \gets f(c_t,a_{t+1})$\;
    $(\text{boxes}_{t+1},\text{scores}_{t+1}) \gets D(c_{t+1})$\;
    $r_t \gets \textbf{UncertaintyReduction}(\text{scores}_t, \text{scores}_{t+1})$\;
    update $\mathbf{n}_{z_t}$ with the observed successor $z_{t+1}$\;
    $\mathcal{T} \gets \mathcal{T} \cup \{(z_t,z_{t+1},r_t)\}$\;
    $t \gets t+1$\;
  }
  compute $\hat P(\cdot\mid z)$ from Dirichlet\; 
  $\mathcal{D} \gets \mathcal{D} \cup \{(\mathcal{T}, \hat P)\}$\;
  \If{\textbf{ImageStop}$(\mathcal{D},\epsilon_r,\epsilon_P)$}{\textbf{break}}
}

\BlankLine
\textbf{Offline Training:} minimize $\mathcal{L}_{\mathrm{RM}}$ on $\mathcal{D}$ to update $\theta$\;
\textbf{Inference:} replace UCB with RM decision rules (policy-, reward-, or hybrid-driven) to select $z_{t+1}$\;

\end{algorithm}
\subsection{Self-Evolving Loop}
\label{sec:3.5}

\paragraph{Overall Workflow.}
\begin{enumerate}
\item \textbf{Sampling Phase (Bandit):}  
UCB-driven exploration generates multiple weak Markov trajectories $\mathcal{T}_i$ for each image $x_i$, while Dirichlet updates produce the corresponding transition prior $\hat{P}_i$.  
These image-level units $(\mathcal{T}_i, \hat{P}_i)$ are accumulated into the offline buffer $\mathcal{D}$.
\item \textbf{Offline Training:}  
The Reward–Policy Model (RM) is optimized on $\mathcal{D}$ by minimizing $\mathcal{L}_{\mathrm{RM}}$, recovering both transition behavior and weak reward patterns.
\item \textbf{Inference Phase:}  
During deployment, UCB exploration is replaced by RM predictions, enabling the agent to perform self-evolving reasoning without online sampling.
\end{enumerate}

\paragraph{Inference Decision Rules.}
During deployment, the next weak Markov unit is selected as
\begin{equation}
z_{t+1}=
\begin{cases}
\arg\max_{z}\,\pi_\theta(z\mid z_t)\\
\arg\max_{z}\,\hat r_\theta(z)\\
\arg\max_z \bigl[\alpha \log \pi_\theta(z\mid z_t) \\ 
 \hspace{4em} + (1-\alpha)\,\widehat{\mathrm{norm}}(\hat{r}_\theta(z))\bigr], 
\end{cases}
\end{equation}
\noindent\textit{Selection modes:}
(i) policy-driven (default);  
(ii) reward-driven;  
(iii) hybrid, where $\alpha \in [0,1]$ balances the two heads.

\paragraph{Complexity and Practicality.}
The sampling cost scales linearly with trajectory length and action space size. The RM is a compact 3-layer MLP with dual heads (20MB), keeping OVOD-Agent \textbf{LLM-free} and introducing only minor memory overhead during inference, allowing it to be incorporated into different OVOD detectors with minimal modification.
\section{Experiments}
\label{sec:exp}

In this section, we evaluate the performance of \textbf{OVOD-Agent} on the COCO \cite{coco} and LVIS \cite{lvis} benchmarks, conduct ablation studies to analyze its core components, and discuss limitations as well as representative failure cases.

\subsection{Main results}

We evaluate \textbf{OVOD-Agent} on the \textbf{COCO} and \textbf{LVIS} benchmarks under the open-vocabulary object detection (OVOD) setting by plugging it into four representative base detectors: \textbf{GroundingDINO}, \textbf{YOLO-World}, \textbf{GroundingDINO 1.5} (API access), and \textbf{DINO-X Pro} (API access).
The overall results are summarized in Table \ref{tab2}.

\paragraph{Datasets.}
Open-vocabulary detectors typically struggle on \textit{rare categories}, which are heavily underrepresented in existing training corpora. 
To evaluate both general and long-tailed performance, we adopt the COCO and LVIS benchmarks under the standard open-vocabulary detection setting. 
For LVIS, we report results on the full LVIS val split (20k images) and the widely used LVIS minival subset (5k images). 
Following prior works such as Detic \cite{Detic} and GLIP \cite{glip}, LVIS minival is formed by selecting the first 5k images in the official validation index, providing a fast yet comparable protocol for OVD evaluation.

\paragraph{Results analysis.} 
As shown in Table~\ref{tab2}, \textbf{OVOD-Agent} provides moderate and consistent
improvements across all base detectors. On LVIS val, the rare-category metric AP$_r$ improves by \textbf{+2.7}, \textbf{+2.4}, \textbf{+1.4}, and \textbf{+1.2} for GroundingDINO, YOLO-World, GroundingDINO 1.5, and DINO-X Pro, respectively, demonstrating the agent's
effectiveness in long-tailed recognition. These improvements remain consistent
on the LVIS minival subset, where OVOD-Agent increases AP$_r$ by \textbf{+1.6}, \textbf{+1.8}, \textbf{+1.3}, and \textbf{+1.1}. In all cases, the gains in overall AP are steady (ranging from \textbf{+0.5 to +1.2}), indicating that the method enhances rare categories
without negatively affecting common or frequent ones. On COCO2017 val, where
categories are more balanced, OVOD-Agent yields mild improvements of \textbf{+0.6--1.3} mAP, suggesting that the enhanced reasoning mechanism provides
limited but stable benefits even for well-represented classes.
Beyond accuracy, Table~\ref{tab2} also reports $\Delta$Latency, which measures the
average per-image increase in inference time when the agent is integrated. Since
each reasoning step adds one additional detector forward pass, the overhead grows
approximately linearly with the trajectory length. This extra cost remains within
an acceptable range while delivering consistent gains in rare-category accuracy,
reflecting a favorable accuracy--efficiency trade-off.

\begin{table*}[ht]
    \caption{\textbf{Main results on LVIS and COCO benchmarks.} OVOD-Agent improves rare-category detection (AP$_r$) while maintaining stable overall accuracy (AP), with only a small increase in inference latency.}
    \label{tab2}
    \centering
    \footnotesize
    \setlength{\tabcolsep}{4pt}
    \resizebox{\textwidth}{!}{
    \begin{tabular}{lcccccc|cccc|cc|c}
    \toprule
    \multirow{2}{*}{\textbf{Method}} &
    \multicolumn{6}{c|}{\textbf{LVIS$^{val}$}} &
    \multicolumn{4}{c|}{\textbf{LVIS$^{minival}$}} &
    \multicolumn{2}{c|}{\textbf{COCO$^{2017val}$}} &
    \multirow{2}{*}{$\mathbf{\Delta}$\textbf{Latency (ms)}} \\ 
    \cmidrule(lr){2-7}\cmidrule(lr){8-11}\cmidrule(lr){12-13}
    & AP$_r$ & AP$_c$ & AP$_f$ & AP$_{all}$ & $\Delta$AP$_r$ & $\Delta$AP 
    & AP$_r$ & AP$_c$ & AP$_f$ & $\Delta$AP$_r$ & AP$_{all}$ & mAP &  \\
    \midrule           
    GroundingDINO \cite{gdino} & 30.2 & 47.8 & 53.2 & 43.7 & -- & -- 
                  & 35.4 & 51.3 & 55.7 & -- & 52.1 & 52.8 & -- \\
    \rowcolor{gray!10}
    \textbf{+ OVOD-Agent} & 32.9 & 48.5 & 53.6 & 44.4 & \textbf{+2.7} & \textbf{+0.7} 
                 & 37.0 & 52.1 & 56.3 & \textbf{+1.6} & 52.7 & 54.1 
                 & +120 \\
    \midrule
    YOLO-World \cite{yoloworld} & 22.8 & 32.3 & 36.2 & 33.3 & -- & -- 
               & 27.6 & 34.1 & 38.0 & -- & 35.4 & 45.0 & -- \\
    \rowcolor{gray!10}
    \textbf{+ OVOD-Agent} & 25.2 & 33.0 & 36.5 & 33.8 & \textbf{+2.4} & \textbf{+0.5}
                 & 29.4 & 35.0 & 38.2 & \textbf{+1.8} & 35.9 & 45.9 
                 & +90 \\
    \midrule
    GroundingDINO1.5 \cite{gdino1.5} & 42.7 & 48.6 & 52.8 & 49.6 & -- & -- 
               & 48.3 & 50.1 & 54.2 & -- & 56.8 & 58.0 & -- \\
    \rowcolor{gray!10} 
    \textbf{+ OVOD-Agent} & 44.1 & 49.5 & 53.4 & 50.8 & \textbf{+1.4} & \textbf{+1.2} 
                    & 49.6 & 50.9 & 54.7 & \textbf{+1.3} & 57.4 & 58.8 & +145 \\
    \midrule
    DINO-X Pro \cite{dinox} & 48.0 & 52.9 & 56.3 & 53.6 & -- & -- 
               & 52.5 & 54.2 & 57.5 & -- & 60.2 & 61.5 & -- \\
    \rowcolor{gray!10} 
    \textbf{+ OVOD-Agent} & \textbf{49.2} & \textbf{53.5} & \textbf{56.9} & \textbf{54.5} & \textbf{+1.2} & \textbf{+0.9} 
                    & \textbf{53.6} & \textbf{54.8} & \textbf{57.9} & \textbf{+1.1} & \textbf{60.6} & \textbf{62.1} & +155 \\
    \bottomrule
    \end{tabular}
    }
    \vspace{-2mm}
\end{table*}

\subsection{Ablation Study}

We conduct a series of ablation experiments to analyze the internal mechanisms of \textbf{OVOD-Agent}.
All ablations are conducted on the \textbf{LVIS minival} using \textbf{GroundingDINO} as the representative base detector unless otherwise specified.

The ablation design aligns with the overall pipeline of OVOD-Agent and reflects its three-stage process:
(1) the \textit{sampling mechanism}, examining the efficiency of the UCB-based exploration strategy;
(2) the \textit{learning mechanism}, analyzing the effect of explicit Markov-state (Markov–Bandit) modeling in reward optimization; and
(3) the \textit{reasoning mechanism}, evaluating the contribution of Visual-CoT actions and visual priors.
This organization mirrors the agent’s operational flow, providing a systematic validation of each component’s role and contribution.

\subsubsection{Effect of UCB Exploration}

We compare the proposed UCB-based exploration policy with three standard baselines—Random, Greedy-Q, and $\epsilon$-Greedy ($\epsilon{=}0.1$) under a unified weak-MDP formulation. 
All methods follow the same \textit{convergence-based stopping criteria} defined in Sec.~\ref{sec:3.3}, ensuring fair comparison with equal termination conditions for both trajectory- and image-level processes.

\paragraph{Evaluation metrics.} 
To assess exploration quality, we report two quantitative indicators:
(i) \textbf{Top-K@Stop}, the mean reward of the top $K{=}\lfloor0.1n_x\rfloor$ trajectories (with $n_x$ denoting the number of sampled trajectories for image $x$) before convergence, which reflects the quality of high-reward trajectories discovered by the policy;
(ii) \textbf{Pareto-Win Rate (PWR)}, the percentage of images where a policy achieves higher Top-K with an equal or smaller sampling budget, reflecting exploration efficiency.
Higher values indicate stronger exploration.and (iii) \textbf{AI/Human Score}, where GPT-5 \cite{gpt5} and human annotators jointly evaluate trajectory coherence to capture both model-based consistency and human-perceived reasoning quality.

\begin{table}[ht]
    \caption{Comparison of exploration strategies under the unified stopping protocol. Higher is better for all metrics. \textbf{AI} scores are based on a blind GPT-5 evaluation (strategy names anonymized) to ensure fairness and prevent prior knowledge bias.}
    \label{tab3}
    \small
    \setlength{\tabcolsep}{3.7pt} 
    \renewcommand{\arraystretch}{1.05}
    \begin{tabular}{lcccc} 
    \toprule
    \textbf{Strategy} & \textbf{Top-K@Stop} & \textbf{PWR(\%)} & \textbf{AI (Blind)} & \textbf{Human} \\
    \midrule
    Random            & 0.54$\pm$0.02 & 19.1 & 3.0$\pm$0.2 & 2.7$\pm$0.3 \\
    Greedy-Q          & 0.59$\pm$0.02 & 29.7 & 3.3$\pm$0.2 & 3.1$\pm$0.3 \\
    $\epsilon$-Greedy & 0.62$\pm$0.01 & 36.5 & 3.5$\pm$0.1 & 3.3$\pm$0.2 \\
    \rowcolor{gray!10}
    \textbf{UCB (Ours)} & \textbf{0.66}$\pm$0.01 & \textbf{44.8} & \textbf{4.7}$\pm$0.1 & \textbf{4.5}$\pm$0.2 \\
    \bottomrule
    \end{tabular}
    \vspace{-2mm}
\end{table}

According to Table~\ref{tab3}, the UCB policy achieves the highest Top-K@Stop (\textbf{0.66}) and PWR (\textbf{44.8\%}), consistently discovering more high-reward trajectories under the same stopping criteria. 
It also obtains higher \textbf{AI (Blind)} and Human evaluation scores (\textbf{4.7} and \textbf{4.5}), where the AI scores are collected through an anonymized protocol to eliminate potential model priors toward known algorithms. 
These results indicate that the sampled trajectories exhibit clearer semantic progression and are judged as more coherent and meaningful, with the advantages arising purely from improved exploration behavior rather than increased sampling.

\subsubsection{Impact of Markov-State Modeling in RM Training}

We compare two reward model (RM) optimization schemes:  
(1) a \textit{trajectory-only} baseline trained purely from sequential samples, and  
(2) the full \textit{Markov--Bandit} variant that explicitly incorporates the empirical transition matrix $\hat{P}(a'|a,x)$ together with a KL-based transition regularization.  
This design enforces transition-consistent updates, mitigating unstable reward propagation and overfitting to local action modes.

All models are evaluated using four metrics:  
RM loss stability, action entropy, and downstream detection accuracy (AP and AP$_r$).  
Action entropy reflects the diversity and uncertainty of the learned policy—higher entropy indicates richer exploration, while lower values suggest overconfident or mode-collapsed behavior.

\begin{table}[ht]
    \caption{Ablation on explicit Markov-state modeling in reward optimization. Metrics include RM loss stability, action entropy, and detection performance (AP and AP$_r$) measured on the LVIS\textsuperscript{minival} benchmark.}
    \label{tab4}
    \centering
    \normalsize
    \setlength{\tabcolsep}{4pt}
    \renewcommand{\arraystretch}{1.05}
    \resizebox{\linewidth}{!}{
    \begin{tabular}{lcccc}
    \toprule
    \textbf{Training Scheme} & \textbf{RM Loss Std} & \textbf{Action Entropy} & \textbf{AP} & \textbf{AP$_r$} \\
    \midrule
    w/o KL Reg. & 0.037 & 1.41 & 38.2 & 19.0 \\
    \rowcolor{gray!10}
    \textbf{w/ KL Reg. (Full)} & \textbf{0.028} & \textbf{1.55} & \textbf{39.4} & \textbf{20.3} \\
    \bottomrule
    \end{tabular}
    }
    \vspace{-2mm}
\end{table}

Results in Table~\ref{tab4} show that incorporating the KL-based transition regularization reduces RM loss variance from \textbf{0.037} to \textbf{0.028}, indicating more stable reward training. It also increases action entropy (1.41 $\rightarrow$ \textbf{1.55}), suggesting that the transition prior prevents the policy from collapsing into local modes and encourages more balanced exploration. These improvements translate into consistent downstream gains on LVIS minival, with AP rising from 38.2 $\rightarrow$ \textbf{39.4} and AP$_r$ from 19.0 $\rightarrow$ \textbf{20.3}. This validates that explicit Markov-state modeling, acting as a structural regularizer, effectively stabilizes reward learning and contributes meaningfully to final detection performance.

\begin{figure*}[ht]
\centering
\includegraphics[width=1.0\textwidth]{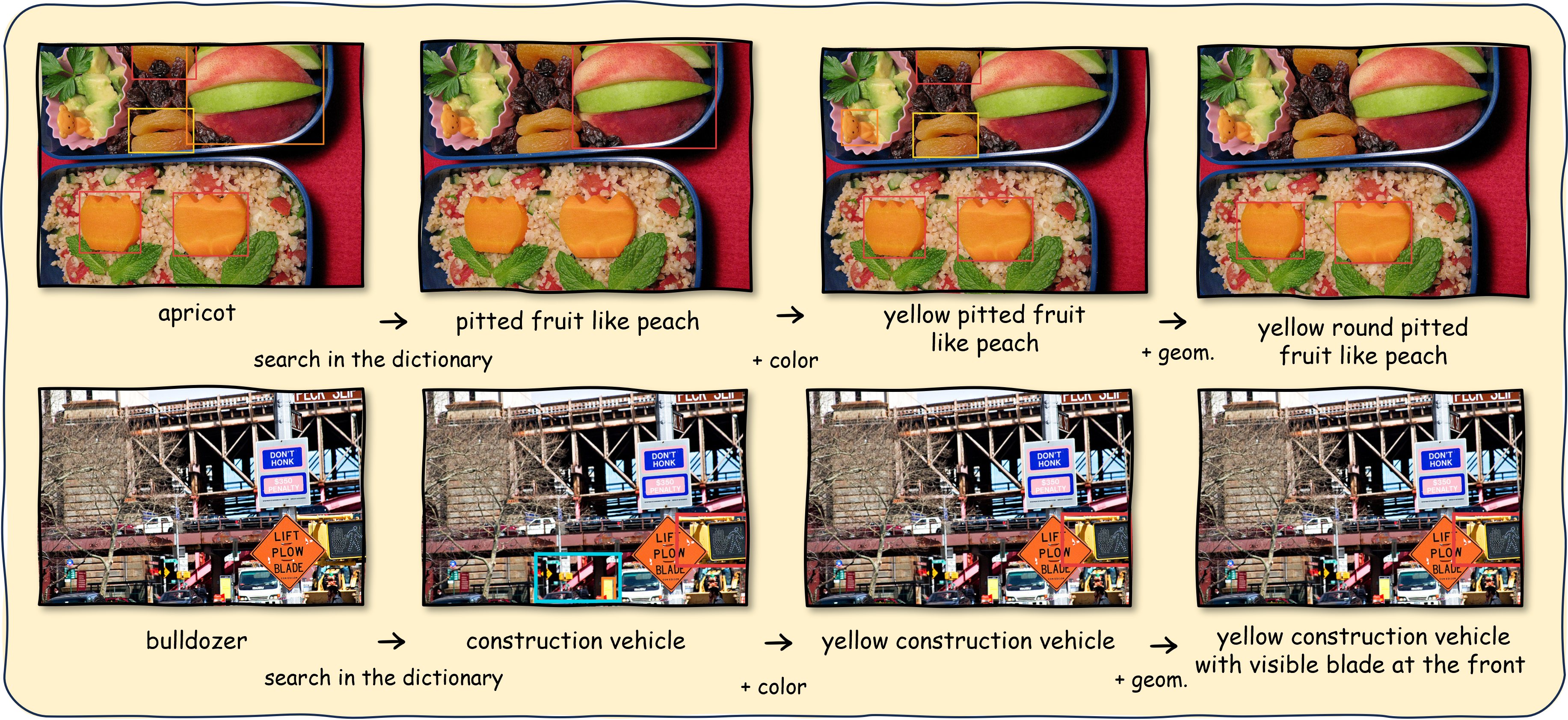}
\caption{\textbf{Failure cases of OVOD-Agent.} Representative examples where the agent fails to correctly identify rare or occluded objects.}
\label{fig3}
\end{figure*}

\subsubsection{Contribution of Visual-CoT Actions and Priors}

We progressively expand the agent's action space from textual reasoning (a$_1$) to the full Visual-CoT set (a$_1$–a$_7$), incorporating attribute- and geometry-aware cues like color, texture, material, lighting, and spatial priors. All experiments use the LVIS minival split with GroundingDINO as the base detector.

As shown in Table~\ref{tab5}, attribute-level Visual-CoT actions lead to consistent performance gains.  
Introducing only the dictionary-based textual action (+a$_1$) yields a moderate improvement, raising AP$_r$ from 35.4 to 36.5.  
When the full action space (+a$_1$–a$_7$) is enabled, rare-category performance further increases to \textbf{37.7} AP$_r$, accompanied by similar gains in AP$_c$, AP$_f$, and overall AP.  
These results demonstrate that structured Visual-CoT actions provide richer semantic refinements and more discriminative attribute cues, thereby enhancing open-vocabulary generalization.

\begin{table}[t]
    \centering
    \caption{
    Ablation on Visual-CoT actions and visual priors.
    Metrics are evaluated on the LVIS minival dataset.
    }
    \label{tab5}
    \normalsize 
    \setlength{\tabcolsep}{8pt} 
    \renewcommand{\arraystretch}{1.05}
    \begin{tabular}{lcccc}
      \toprule
      \textbf{Action Set} & \textbf{AP$_r$} & \textbf{AP$_c$} & \textbf{AP$_f$} & \textbf{AP} \\
      \midrule
      Baseline & 35.4 & 51.3 & 55.7 & 52.2 \\
      +a$_1$ & 36.5 & 52.1 & 55.9 & 53.6 \\
      \rowcolor{gray!10}
      \textbf{+a$_1$–a$_7$} & \textbf{37.7} & \textbf{52.8} & \textbf{56.3} & \textbf{54.5} \\
      \bottomrule
    \end{tabular}
    \vspace{-2mm}
\end{table}

\subsection{Limitations and Failure Analysis}

Despite improvements in rare-category detection, \textbf{OVOD-Agent} faces challenges in extreme long-tail scenarios. Figure~\ref{fig3} illustrates two primary failure modes.

\paragraph{Visual–Semantic Degradation.}
When objects appear in non-canonical states (e.g., \textit{dried apricot}), their appearance deviates significantly from the detector's visual priors. This mismatch causes the reasoning process to over-rely on linguistic priors rather than adapting to degraded visual cues. Sparse transition statistics for such rare states further destabilize reward updates.
\paragraph{Tiny Objects and Clutter.}
For small, occluded objects in cluttered environments (e.g., \textit{bulldozer}), geometric (a$_5$) and spatial (a$_7$) actions often yield noisy reward signals. Consequently, the policy tends to fall back on dictionary lookups (a$_1$) without improving localization. Background clutter can also induce misleadingly high alignment scores with related categories, confusing the reward model.
\paragraph{Discussion.}
These cases highlight two limitations: a sensitivity to semantic–visual mismatches for non-canonical forms and difficulty localizing tiny or occluded rare objects in complex contexts. Future work may require stronger visual priors or adaptive OOD reasoning strategies.

\section{Conclusions}

We present OVOD-Agent, a lightweight Markov-Bandit framework transforming open-vocabulary detection from static matching into proactive visual reasoning. By grounding its Visual-CoT in a discrete, weakly Markovian state distribution and uncertainty-aware exploration, it consistently improves performance across diverse backbones, especially on rare categories. It provides a scalable foundation for self-evolving visual reasoning in open-world settings.


{
    \small
    \bibliographystyle{ieeenat_fullname}
    \bibliography{main}
}

\clearpage
\setcounter{page}{1}
\maketitlesupplementary

\begin{algorithm*}[b]
\caption{Visual Actions $a_1$--$a_7$}
\label{alg:actions}
\DontPrintSemicolon
\SetAlgoLined
\SetKwProg{Fn}{Function}{:}{end}

\begin{multicols}{2}

    \Fn{\textnormal{$a_1$: Dictionary}}{
        obj $\leftarrow$ \textnormal{PARSE\_NOUN}(c)\;
        syn $\leftarrow$ \textnormal{WORDNET\_SYN}(obj)\;
        hyp $\leftarrow$ \textnormal{WORDNET\_HYPER}(obj)\;
        cand $\leftarrow$ syn $\cup$ hyp\;
        tokens $\leftarrow$ \textnormal{FILTER\_VISUAL\_TERMS}(cand)\;
        phrase $\leftarrow$ \textnormal{FORMAT}(``a {} object'', tokens)\;
    }

    \Fn{\textnormal{$a_2$: Color}}{
        hsv $\leftarrow$ \textnormal{TO\_HSV}(r)\;
        clst $\leftarrow$ \textnormal{KMEANS}(hsv, 3)\;
        dom $\leftarrow$ \textnormal{LARGEST\_CLUSTER}(clst)\;
        col $\leftarrow$ \textnormal{HSV\_TO\_COLOR}(dom)\;
        phrase $\leftarrow$ \textnormal{FORMAT}(``{} color'', col)\;
    }

    \Fn{\textnormal{$a_3$: Texture}}{
        lbp $\leftarrow$ \textnormal{LBP}(r)\;
        glcm $\leftarrow$ \textnormal{GLCM}(r)\;
        feats $\leftarrow$ \{lbp, glcm\}\;
        tex $\leftarrow$ \textnormal{MATCH\_TEXTURE}(feats)\;
        phrase $\leftarrow$ \textnormal{FORMAT}(``{} texture'', tex)\;
    }


    \Fn{\textnormal{$a_4$: Background}}{
        fg $\leftarrow$ \textnormal{FG\_MASK}(r)\;
        bg $\leftarrow$ r - fg\;
        clt $\leftarrow$ \textnormal{BG\_CLUTTER}(bg)\;
        tag $\leftarrow$ \textnormal{IF}(clt > $\tau$, ``cluttered'', ``clean background'')\;
        phrase $\leftarrow$ \textnormal{FORMAT}(``object against {}'', tag)\;
    }

    \vfill\null\columnbreak

    \Fn{\textnormal{$a_5$: Geometry}}{
        ar $\leftarrow$ \textnormal{ASPECT\_RATIO}(r)\;
        sc $\leftarrow$ \textnormal{BBOX\_SCALE}(r)\;
        geom $\leftarrow$ \textnormal{CLASSIFY\_GEOM}(ar, sc)\;
        phrase $\leftarrow$ \textnormal{FORMAT}(``{} shaped'', geom)\;
    }

    \Fn{\textnormal{$a_6$: Lighting}}{
        hist $\leftarrow$ \textnormal{V\_CHANNEL\_HIST}(r)\;
        mean $\leftarrow$ \textnormal{MEAN}(hist)\;
        var  $\leftarrow$ \textnormal{VAR}(hist)\;
        
        cond $\leftarrow$ \textnormal{IF3}(
            $mean < \tau_{\text{dark}}$, ``underexposed'',\;
            \hspace{2cm} $mean > \tau_{\text{bright}}$, ``overexposed'',\;
            \hspace{2cm} $var > \tau_{\text{shadow}}$, ``shadowed'',\;
            \hspace{2cm} ``well-lit'' )\;

        phrase $\leftarrow$ \textnormal{FORMAT}(``{} lighting'', cond)\;
    }

    \Fn{\textnormal{$a_7$: Spatial}}{
        rel $\leftarrow$ \textnormal{SPATIAL\_REL}(r, layout)\;
        phrase $\leftarrow$ \textnormal{FORMAT}(``the {} object'', rel)\;
    }

\end{multicols}

\end{algorithm*}

\section{Appendix}
\label{sec:appendix}
This appendix provides additional technical details that complement the main paper. We first present the full set of visual-action operators used by \textbf{OVOD-Agent}. Next, we provide an expanded case study that demonstrates how the agent incrementally refines its textual hypotheses using both low-level and high-level visual cues. We further include a comparison between OVOD-Agent and other LLM-based COT methods, emphasizing the differences in inference latency across these approaches. Finally, we provide the exact GPT-5 evaluation prompts and scoring rubric used to assess trajectory coherence and groundedness, ensuring transparency and reproducibility of our analysis.

\subsection{Visual Actions Space}

This section presents the pseudocode for the seven interpretable visual actions
($a_1$--$a_7$) used by \textbf{OVOD-Agent}, as summarized in
Algorithm~\ref{alg:actions}. Each action extracts a specific visual cue from the
ROI (e.g. color, texture, geometry, background, lighting, or spatial relation)
and maps it to a short linguistic attribute that is later used to update the
evolving caption in the main reasoning algorithm.

To ensure reproducibility, all visual cues are computed using standard computer
vision toolkits. Specifically, we employ \textbf{OpenCV} for RGB--HSV color
conversion, K-means clustering, edge detection, and basic shape analysis;
\textbf{scikit-image} for LBP/GLCM texture extraction, brightness histogram
estimation, and foreground/background masking; and \textbf{NumPy/SciPy} for
region-level statistics, histogram aggregation, and geometric feature
computation.

These operations enable OVOD-Agent to extract stable and interpretable visual
cues directly from the image, providing a consistent basis for subsequent
textual refinement.

\begin{table*}[htb]
\caption{\textbf{Comparison with LLM-guided reasoning modules.} 
OVOD-Agent achieves competitive rare-category improvements while keeping inference in the \emph{millisecond} regime, 
whereas LLM-based online reasoning methods (e.g., RALF) require \emph{second-level} latency.}
\label{tab:llm_comparison}
\centering
\footnotesize
\setlength{\tabcolsep}{8pt}
\resizebox{\textwidth}{!}{
\begin{tabular}{lcccc|c|c|c|c}
\toprule
\textbf{Method} & 
\textbf{LVIS AP$_r$} & \textbf{AP$_c$} & \textbf{AP$_f$} & \textbf{AP$_{all}$} & 
\textbf{COCO AP$^{N}_{50}$} & 
\textbf{Avg Latency} & 
\textbf{Worst-case} & 
\textbf{LLM Usage} \\
\midrule

\textbf{GroundingDINO (baseline)} \cite{gdino} & 
35.4 & 51.3 & 55.7 & 52.1 & 
30.8 & 
\textbf{25 ms} & 
25 ms & 
\textbf{Free} \\

\textbf{RALF (LLM-based RAG)} \cite{ralf} & 
38.6 & 52.0 & 56.1 & 52.9 & 
33.2 & 
\textbf{1.5 s} & 
3.0 s & 
\textbf{Online} \\

\textbf{CoT-PL (Visual CoT)} \cite{cotpl} & 
37.4 & 51.8 & 55.9 & 52.7 & 
32.5 & 
\textbf{1.2 s} & 
2.5 s & 
\textbf{Offline} \\

\textbf{DVDet (VQA-refined descriptors)} \cite{llmguided2} & 
36.2 & 51.0 & 55.3 & 52.0 & 
31.4 & 
\textbf{30 ms} & 
45 ms & 
\textbf{Offline} \\

\textbf{LLMDet} \cite{llmdet} & 
40.8 & 43.1 & 54.3 & 48.3 & 
55.6 & 
35 ms & 
50 ms & 
\textbf{Offline} \\

\rowcolor{gray!10}
\textbf{OVOD-Agent (Ours)} & 
\textbf{37.0} & \textbf{52.1} & \textbf{56.3} & \textbf{52.7} & 
\textbf{33.4} & 
\textbf{55 ms} & 
\textbf{175 ms} & 
\textbf{Free} \\

\bottomrule
\end{tabular}
}
\begin{flushleft}
\end{flushleft}
\vspace{-1mm}
\end{table*}

\subsection{Detailed Step-by-Step Case Study}
\begin{figure*}[ht]
\centering
\vspace{-50pt}
\includegraphics[width=0.80\textwidth]{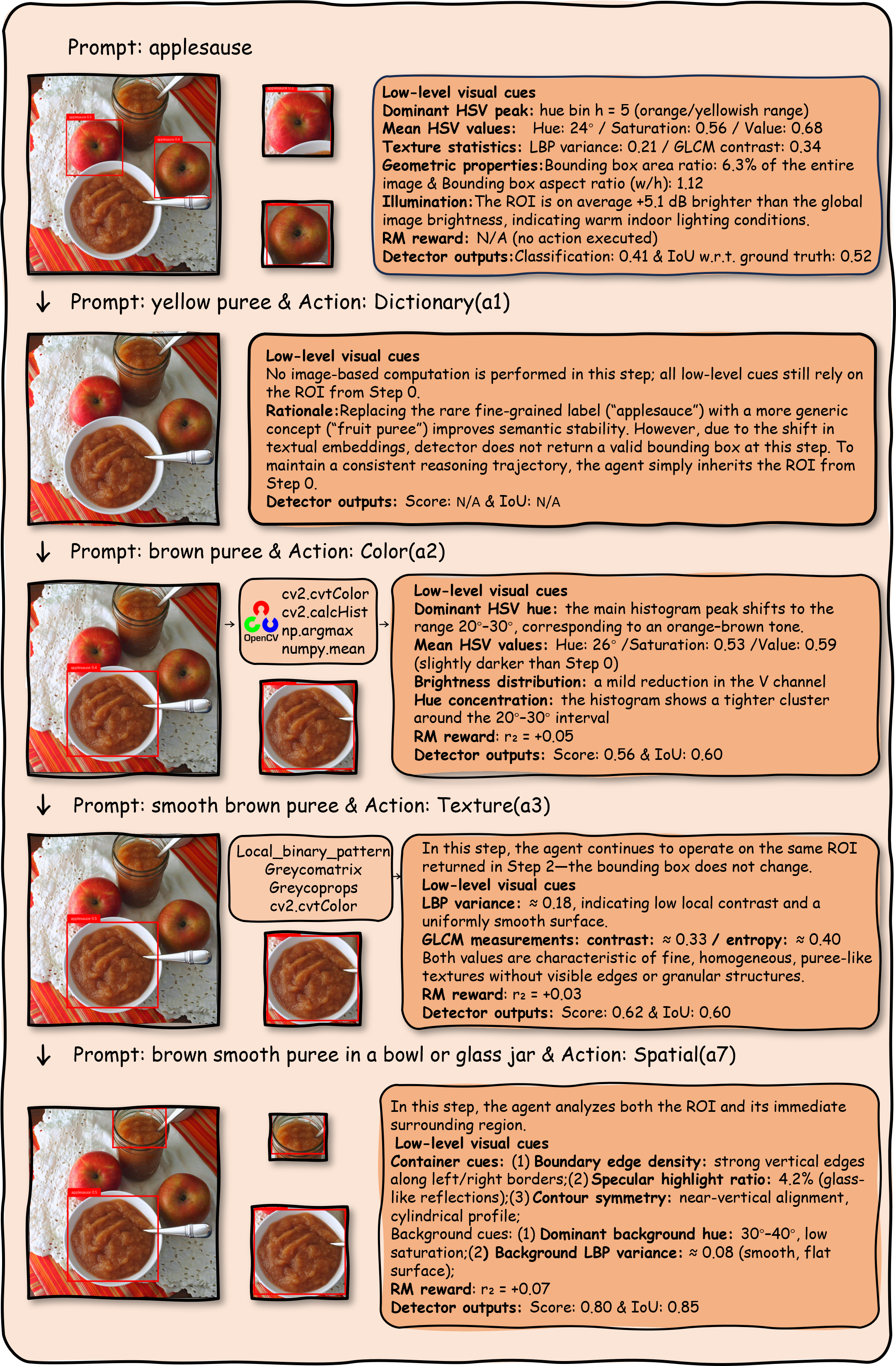} 
\caption{
Step-by-step Case Study of \textbf{OVOD-Agent}, showing how visual actions 
(color, texture, container, background, spatial cues) progressively refine 
the caption and stabilize detector grounding.
}
\label{fig:case-study}
\end{figure*}

 To illustrate how \textbf{OVOD-Agent} performs multi-step visual reasoning,
we present a detailed case study (Fig.~\ref{fig:case-study}) that traces the
complete prediction trajectory from an initial noun-only caption to a fully
grounded, attribute-rich description. At each reasoning step, the agent executes
one visual action, extracts a specific cue from the ROI (e.g. color via HSV
analysis, texture via LBP/GLCM, or high-level cues from container/background
geometry), converts the cue into a linguistic attribute, and updates the
slot-based caption accordingly. For each step, we report: (1) the extracted
visual evidence, (2) the updated caption, (3) the reward produced by the RM,
and (4) the detector's grounding response (score and IoU). This case study
demonstrates how progressive, attribute-aware refinement enables
\textbf{OVOD-Agent} to stabilize open-vocabulary grounding even when initial
predictions are incomplete or the detector temporarily fails to produce a
bounding box.

\subsection{Comparison with LLM-guided Methods}
To demonstrate the efficiency of our \textbf{LLM-Free} paradigm, this section contrasts OVOD-Agent with representative LLM-guided modules, including RALF, CoT-PL, DVDet, and LLMDet. As summarized in Table~\ref{tab:llm_comparison}, our approach eliminates the heavy dependencies that plague existing methods.

\paragraph{Inference Latency Bottleneck.} 
Online reasoning methods like \textbf{RALF} are severely limited by their reliance on real-time LLM calls. Each ``detection $\to$ LLM $\to$ re-detection'' cycle drags the latency into the \emph{second-level} regime (1.5 s), making them impractical for real-time deployment. While \textbf{CoT-PL}, \textbf{DVDet}, and \textbf{LLMDet} attempt to achieve faster inference (30--35 ms), they simply shift the burden to the training phase. These offline methods require massive computational resources and time to generate millions of pseudo-labels or descriptors using heavy LLMs (e.g., Qwen2-72B) before training can even begin.

\paragraph{The Superiority of LLM-Free Reasoning.} 
In sharp contrast, \textbf{OVOD-Agent} is the only framework that remains entirely \textbf{LLM-Free} across both training and inference. It replaces expensive linguistic reasoning with lightweight visual actions (color, texture, geometry, spatial cues). Despite the lack of LLM intervention, OVOD-Agent achieves a competitive 37.0 $AP_r$ on LVIS, outperforming several methods that rely on VQA-refined descriptors (e.g., DVDet at 36.2 $AP_r$). By formulating reasoning as a Markov-Bandit process, we achieve roughly $2.2 \times$ faster inference than the base detector with reasoning, maintaining a strict millisecond latency (55 ms).

\subsection{Blind GPT-5 Trajectory Scoring}

\begin{figure*}[ht]
\centering
\vspace{-30pt}
\includegraphics[width=\textwidth]{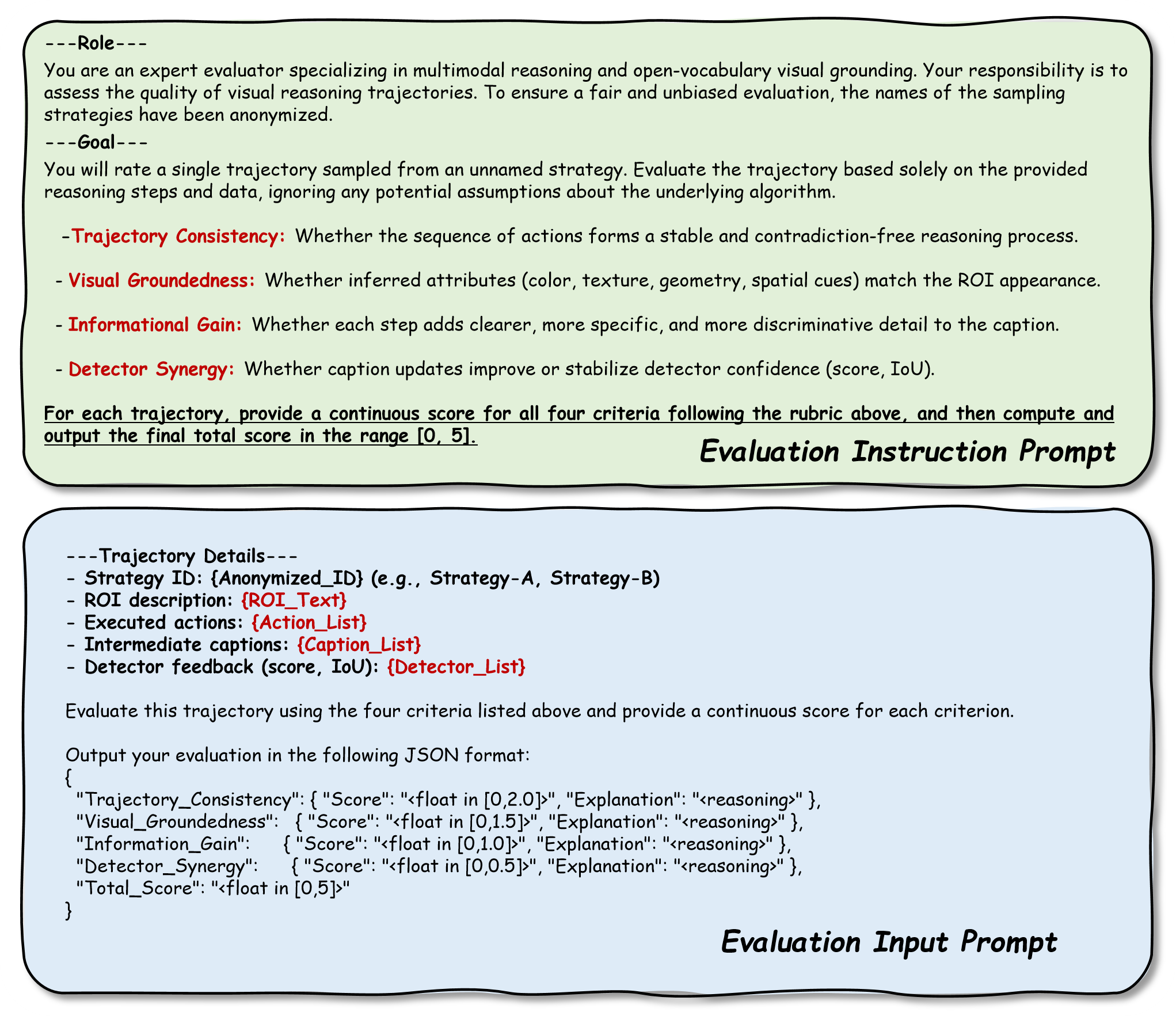}
\caption{
Evaluation protocol for \textbf{blind} GPT-5 trajectory scoring, including the instruction prompt defining the evaluator's role and the \textbf{anonymized} input prompt to ensure unbiased assessment.
}
\label{fig:gpt-prompt}
\end{figure*}

For completeness, we include the prompt template used for the \textbf{blind GPT-5 evaluation}. GPT-5 does not participate in inference; it is used only to assign a continuous weak score to each sampled trajectory as an offline evaluator. To eliminate potential bias toward well-known algorithms, we implemented an \textbf{anonymized protocol} where all strategy names were replaced with generic identifiers (e.g., Strategy-A). As shown in Fig.~\ref{fig:gpt-prompt}, the evaluation consists of an instruction prompt (defining the evaluator's role and the \textbf{anonymization requirement}) and an input prompt containing the trajectory details. GPT-5 rates each trajectory according to the four criteria introduced in the main paper and outputs a final aggregated score in the range $[0,5]$ in JSON format.

\end{document}